\documentclass[letterpaper, 10 pt, conference]{ieeeconf}  

\IEEEoverridecommandlockouts                              

\usepackage{graphicx} 
\usepackage{fancyhdr}
\fancypagestyle{plain}{
	\fancyhf{} 
	\fancyfoot[L]{\footnotesize \copyright2023 IEEE. Personal use of this material is permitted. Permission from IEEE must be obtained for all other uses, in any current or future media, including reprinting/republishing this material for advertising or promotional purposes, creating new collective works, for resale or redistribution to servers or lists, or reuse of any copyrighted component of this work in other works.}
	\fancyfoot[C]{}
	\fancyfoot[R]{}

}

\title{\LARGE \bf
	Automatic Generation of Labeled Data for Video-Based Human Pose Analysis via NLP applied to YouTube Subtitles
}

\author{Sebastian Dill$^1$, Susi Zhihan Li$^1$, Maurice Rohr$^1$, Maziar Sharbafi$^2$, and Christoph Hoog Antink$^1$
	\thanks{$^1$KIS*MED (AI Systems in Medicine Lab), Technische Universität Darmstadt, Darmstadt, Germany. $^2$Lauflabor Locomotion Laboratory, Centre for Cognitive Science, Technische Universität Darmstadt, Darmstadt, Germany.
		Corresponding author is Mr. Sebastian Dill, {\tt\small dill@kismed.tu-darmstadt.de}.}%
	\thanks{The authors gratefully acknowledge financial support provided by the Hessian Ministry for Digital Strategy and Development [Hessisches Ministerium für Digitale Strategie und Entwicklung, Distr@l-Förderlinie 2, ``SG4smartmedication'', 21\_0038\_2A]. This study is partly supported by the Hessian Ministry of Higher Education, Science, Research and Art
		and its LOEWE research priority program under the grant ``WhiteBox''. 
	}%
}

\begin{document}
	
	\maketitle
	\thispagestyle{plain}


\begin{abstract}
With recent advancements in computer vision as well as machine learning (ML), video-based at-home exercise evaluation systems have become a popular topic of current research. However, performance depends heavily on the amount of available training data. Since labeled datasets specific to exercising are rare, we propose a method that makes use of the abundance of fitness videos available online. Specifically, we utilize the advantage that videos often not only show the exercises, but also provide language as an additional source of information. With push-ups as an example, we show that through the analysis of subtitle data using natural language processing (NLP), it is possible to create a labeled (irrelevant, relevant correct, relevant incorrect) dataset containing relevant information for pose analysis. In particular, we show that irrelevant clips ($n=332$) have significantly different joint visibility values compared to relevant clips ($n=298$). Inspecting cluster centroids also show different poses for the different classes.
\end{abstract}


\section{INTRODUCTION}
Physical therapy is a crucial step in the treatment of many injuries and diseases. Applications range from acute injuries to rare conditions such as hemophilia, where physiotherapy and rehabilitation can help prevent disabilities and preserve a patient's autonomy~\cite{Heijnen}. While ideally, physical therapy is performed under supervision of a medical professional who can offer individual and immediate feedback, most people do not have the resources to visit a training session regularly. Furthermore, home exercises have been shown to be beneficial to the healing process~\cite{proffitt} even without supervision through an expert. On the other hand, wrong executions, misjudgement of one's fitness level, and overexertion might lead to an inefficient training or even worse, serious injuries~\cite{Jones}. To help mitigate these problems, an automated evaluation system can be applied to assess the quality of exercise execution and lessen the need for human supervision. Building on the advances in computer vision in recent years, significant research has been conducted on video-based human pose estimation and motion capture. One of the most commonly used tools to extract pose data from videos is MediaPipe Pose based on the BlazePose model~\cite{blazepose}.
An example of an ML exercise evaluation system that assess how well an exercise is performed based on posture data extracted from videos alone, is proposed by Liu and Chu~\cite{Liu}. They not only identified the overall correctness of the exercises but also which body part was responsible for the wrong posture. While such a system is promising, it is dependent on the amount of available labeled training data. Since labeled training datasets specific to exercising are rare~\cite{weitz}, we propose to use the almost unlimited resource of videos available online through social video platforms such as YouTube. Videos are recorded by medical professionals, physical therapists, personal trainers, and amateurs, who give exercise advice while also showing positive and negative examples of execution. Besides being very diverse in terms of both video quality and exercise quality, these videos may also have the advantage of a detailed exercise description being available through auto-generated subtitles. Hence, they can contain realistic training data that can be identified by the content of the subtitles.

In this work, we utilize natural language processing (NLP) to automatically analyse the subtitles of YouTube videos. Specifically, we implement a text relevance categorizer to recognize relevant parts of the video and reject \textit{irrelevant} parts, classifying the remaining exercise instances as \textit{correct} and \textit{incorrect}. With that, we are able to automatically create a large amount of clips of relevant video material that show a person performing the intended exercise as well as labeling it with regards to exercise quality for ML training purposes. Lastly, we also employ a text summarizer to describe the incorrect samples and increase the labels' level of detail.

\section{METHODS} \label{section:methods}
\begin{figure*}[!h] 
	\includegraphics[trim={0 7cm 0 7cm},clip,width=1\textwidth]{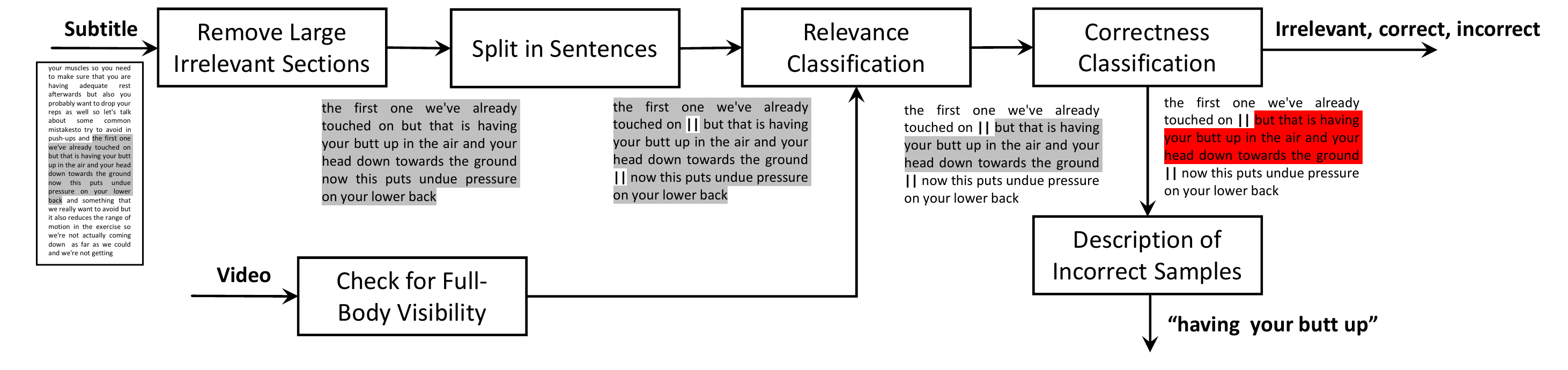}
	\caption{The pipeline to create labeled videos. After removing large irrelevant sections using a first set of keywords, subtitle data is split into sentences. Combining full body visibility and a second set of keywords, sentences are classified in terms of relevance (see Fig.~\ref{fig:sentence_classification}). Relevant sections are then classified as correct or incorrect, while the latter are summarized to identify the type of incorrectness.}
	\label{fig:approach} 
\end{figure*}

In this work, the push-up was selected as an example exercise since it is a prevalent whole-body movement that can be performed without additional equipment. 
The full approach can be seen in Fig.~\ref{fig:approach} and the individual steps are detailed in the following.

\subsection{Video Selection}
A video search was conducted through the YouTube API with the search terms ``learn push up'' (limited to the 30 most relevant results), ``push up mistakes'' (40 results), ``correct OR right push up'' (30 results), ``push up form'' (20 results) and ``perfect push up'' (20 results). These search parameters were optimized to yield the results deemed most relevant by a human observer. After removing videos without English subtitles as well as any duplicates, there were 67 videos left.

\subsection{Relevance Classification through Keywords}
As shown in Fig.~\ref{fig:approach}, in a first step, the subtitle data is processed ``as is'' (i.e. without splitting into sentences) to identify large irrelevant sections in the data using a first set of keywords (KW) and anti-keywords (AKW). Here, both the KW and AKW are defined as two-word expressions about push-ups. The AKW relate to \textit{irrelevant exercises} e.g. they describe common push-up variations in various spellings, such as ``triangle push(-)up(s)''. On the other hand, KW ($n$=64) relate to the standard exercise, e.g. ``ideal push(-)up(s)''. Since it is possible for a video to also feature undesired exercises, other frequent exercise names ($n$=9) such as ``squat(s)'' are also considered to be AKW. Following the occurrence of an AKW, the whole text was marked as \textit{irrelevant} up to the occurrence of a KW.

Next, the remaining text was split into sentences through the NNsplit library\footnote{https://github.com/bminixhofer/nnsplit} utilizing an LSTM-approach. Short sentences with less than 20 characters were excluded and long sentences (in terms of word count and display time) were split.

\begin{figure}[!t] 
	\centerline{\includegraphics[trim={2cm 5cm 11cm 1cm },clip,width=1\columnwidth]{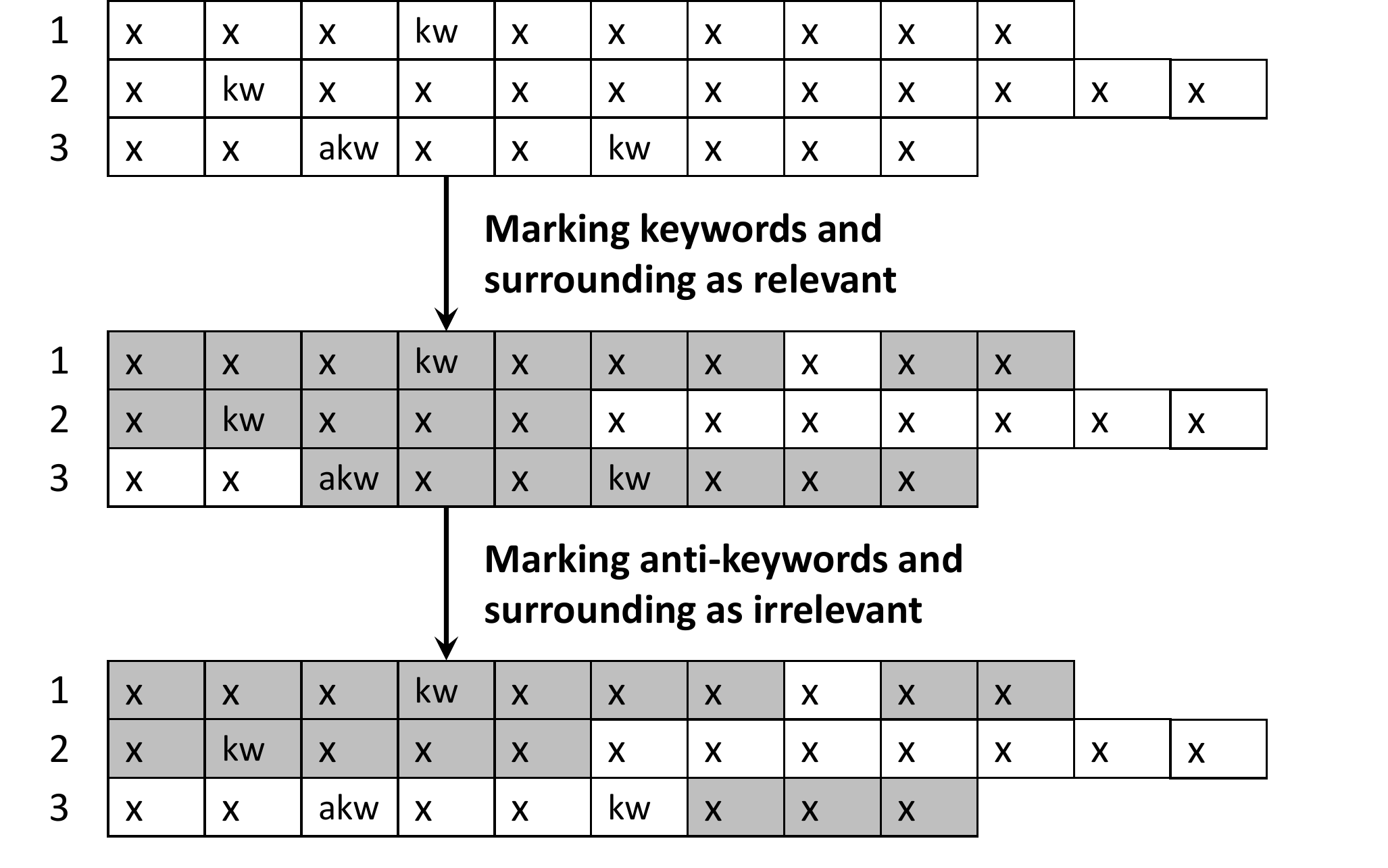}}
	\caption{Sentence Classification with $k=3$ words surrounding the keywords (KW) and anti-keywords (AKW). Here, sentence 1 would be considered \textit{relevant}, sentences 2 and 3 would be considered \textit{irrelevant}, even though sentece 3 contains a KW.}
	\label{fig:sentence_classification}
\end{figure}

A second set of KW and AKW was defined to then classify each sentence as detailed in Fig.~\ref{fig:sentence_classification}. Since other exercises were already filtered out, these (A)KW focus on differentiating whether the speaker is talking about the push-up or about something else, e.g. channel subscription. Hence, the KW ($n$=96) include descriptions of body parts such as ``elbow(s)'' as well as general exercise terms like ``begin'' or ``upward''. The AKW ($n$=15) include terms like ``subscribe'' or ``hello''. When a KW is found, all $k=3$ words surrounding it are marked as \textit{relevant}. Next, all $k=3$ words surround the AKW are marked as \textit{irrelevant}, overwriting previous markings to reduce the chance of wrongly classifying irrelevant parts as relevant. If a sentence contains more \textit{relevant} words than \textit{irrelevant}, the whole sentence is deemed \textit{relevant} and vice versa. 

For this approach to work, an assumption of synchronisation has to be made, meaning that the person is talking about the exercise at the same time as they are showing it. This also includes that \textit{relevant} parts of the video should show the full body at the timestamp of the respective subtitle. To improve the likelihood of this condition, we incorporated video information by checking for full-body visibility through the MediaPipe Pose library.

\subsection{Correctness Classification and Description of Incorrect Samples}
The sentences identified as \textit{relevant} are then classified in terms of \textit{correct} or \textit{incorrect} execution of the push-up. The classification is performed by a 3-gram model trained on 794 sentences, manually labeled as \textit{correct} (551 sentences) or \textit{incorrect} (243 sentences).

Lastly, the sentences classified as \textit{relevant, incorrect} were summarized to produce labels detailing the error. Most text summary methods are centered around reducing the number of sentences. Since the text to be summarized only consists of one sentence at a time, our approach instead focuses on generating key phrases by extracting a span of words from the existing sentences. For the extraction, the approach relies firstly on dependencies between words (e.g. verb and body part) and secondly, if no dependency can be found, on the occurrence of keywords (as described above) and their context. 

\subsection{Video Segmentation and Clustering}
With the presented methods, the 67 videos considered were segmented into 332 \textit{irrelevant} (average length 1672.5 frames), 233 \textit{correct} (175.8 frames) and 65 \textit{incorrect} clips (158.7 frames). Since the ultimate goal is to use the labeled video clips as a dataset for a computer vision-based classifier, we processed all clips with MediaPipe for evaluation. The MediaPipe output consists of x-y-z-coordinates of 33 different pose landmarks as well as an estimate for visibility between 0 and 1 for every landmark. For details on the landmarks, see Table~\ref{tab:pvalues} in the results. As an alternative to manual expert evaluation of the classification results, we propose to use statistical analysis and a k-means clustering algorithm on the extracted pose data to find general differences between the classes identified via NLP.

\section{RESULTS AND DISCUSSION}
For a first analysis, we looked at the visibility values. Out of the clips where all landmarks were recognized, we could observe that the average landmark visibility was lower for clips that were deemed \textit{irrelevant} compared to clips that were considered \textit{relevant}. For every clip and every landmark, we calculated the average visibility and performed the rank-sum test. The resulting difference in group medians and p-values are given in Table~\ref{tab:pvalues}.
\begin{table}[h]
	\caption{Comparison of the median visibility of all landmarks. Ankles, heels and feet indices are less visible for clips deemed irrelevant via NLP.}
	\label{tab:pvalues}
	\begin{center}
		\begin{tabular}{lrr}
			& \multicolumn{2}{c}{$\Delta$ median visibility / p-value}\\
            & \multicolumn{1}{c}{left} & \multicolumn{1}{c}{right}\\
            NOSE & \multicolumn{2}{c}{0 / 0.319}\\
            EYE\_INNER & 0 / 0.424 & 0 / 0.330\\
            EYE & 0 / 0.318 & 0 / 0.262\\
            EYE\_OUTER & 0 / 0.323 & 0 / 0.265\\
            EAR & 0 / 0.431 & 0 / 0.221\\
            MOUTH & 0 / 0.630 & 0 / 0.508\\
            SHOULDER & 0 / 0.469 & 0 / 0.087\\
            ELBOW & -0.059 / 0.340 & -0.033 / 0.966\\
            WRIST & -0.061 / 0.607 & -0.022 / 0.901\\
            PINKY & -0.051 / 0.568 & -0.004 / 0.965\\
            INDEX & -0.052 / 0.535 & -0.005 / 0.830\\
            THUMB & -0.047 / 0.485 & -0.004 / 0.612\\
            HIP & 0 / 0.199 & 0 / 0.381\\
            KNEE & 0.126 / 0.489 & -0.132 / 0.834\\
            ANKLE & \textbf{-0.061 / 0.043} & \textbf{-0.176 / 0.008}\\
            HEEL & \textbf{-0.074 / 0.007} & \textbf{-0.137 / 0.001}\\
            FOOT\_INDEX & \textbf{-0.116 / 0.003} & \textbf{-0.094 / 0.001}\\
		\end{tabular}
	\end{center}
\end{table}
If we assume a p-value of less than 0.05 to indicate statistical significance, we can see that the visibility of the markers ANKLE, HEEL, and FOOT\_INDEX is lower for the \textit{irrelevant} clips in a statistically significant way.

\begin{figure}[!h] 
	\centerline{\includegraphics[width=1\columnwidth]{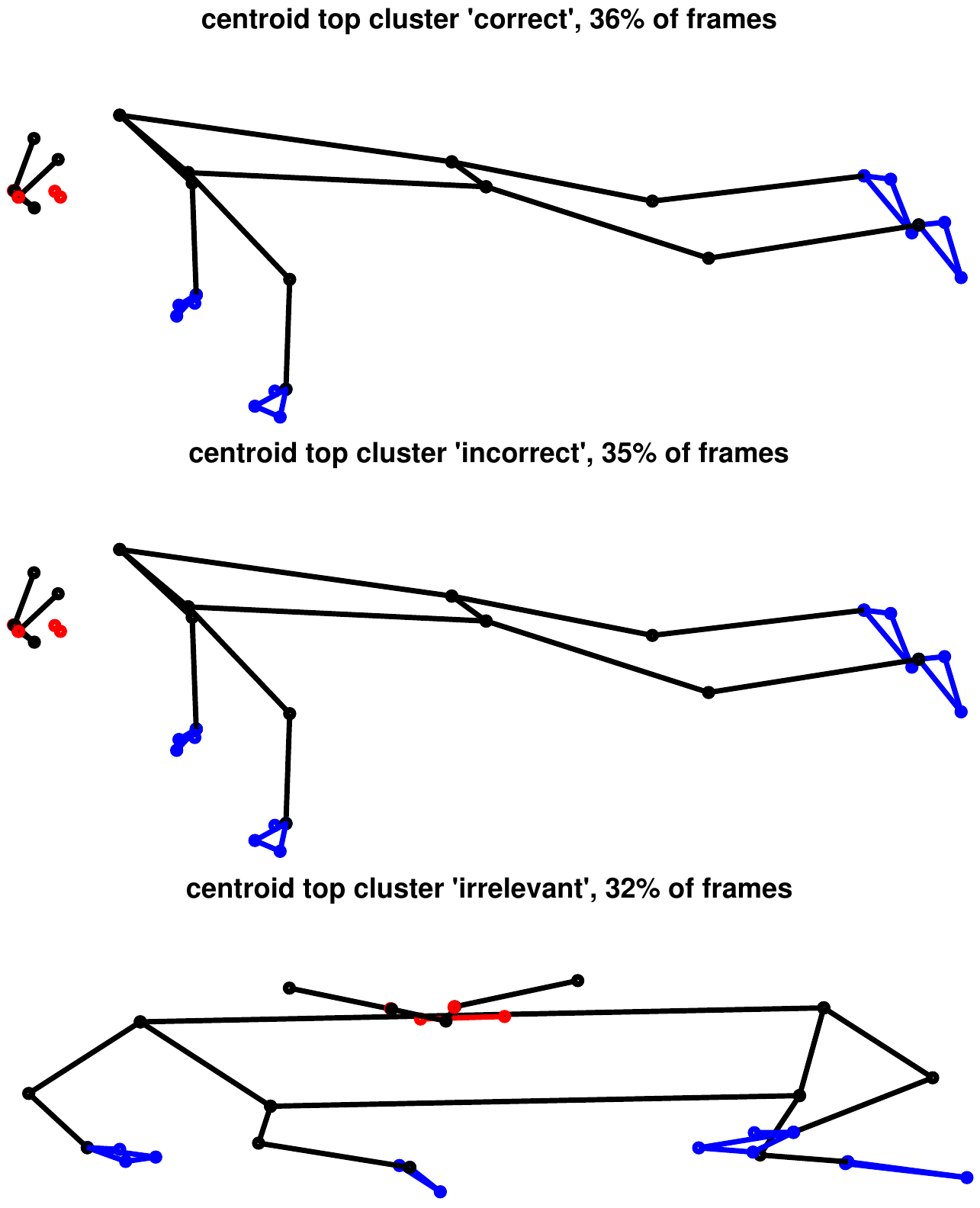}}
	\caption{Visualization of the centroids of the top clusters.  ``relevant, correct'' and ``relevant, incorrect'' predominantly map to the same cluster showing a person in push-up position, while ``irrelevant'' mainly maps to a cluster showing a person facing the viewer.}
	\label{fig:clusters_3} 
\end{figure}

In order to do a qualitative evaluation, we cluster all frames of all clips \emph{independent of their classification}. Next, we identify the top cluster, i.e. the cluster that contains the largest percentage of frames, for each class. The visualization of the respective centroids are shown in Fig.~\ref{fig:clusters_3}.
As can be seen, both \textit{relevant, correct} as well as \textit{relevant, incorrect} show a person in a push up position. In fact, both map to the same cluster. On the other hand, the centroid of the top cluster for \textit{irrelevant} shows a person facing the viewer with distorted legs.

\begin{figure}[!h] 
	\centerline{\includegraphics[width=1\columnwidth]{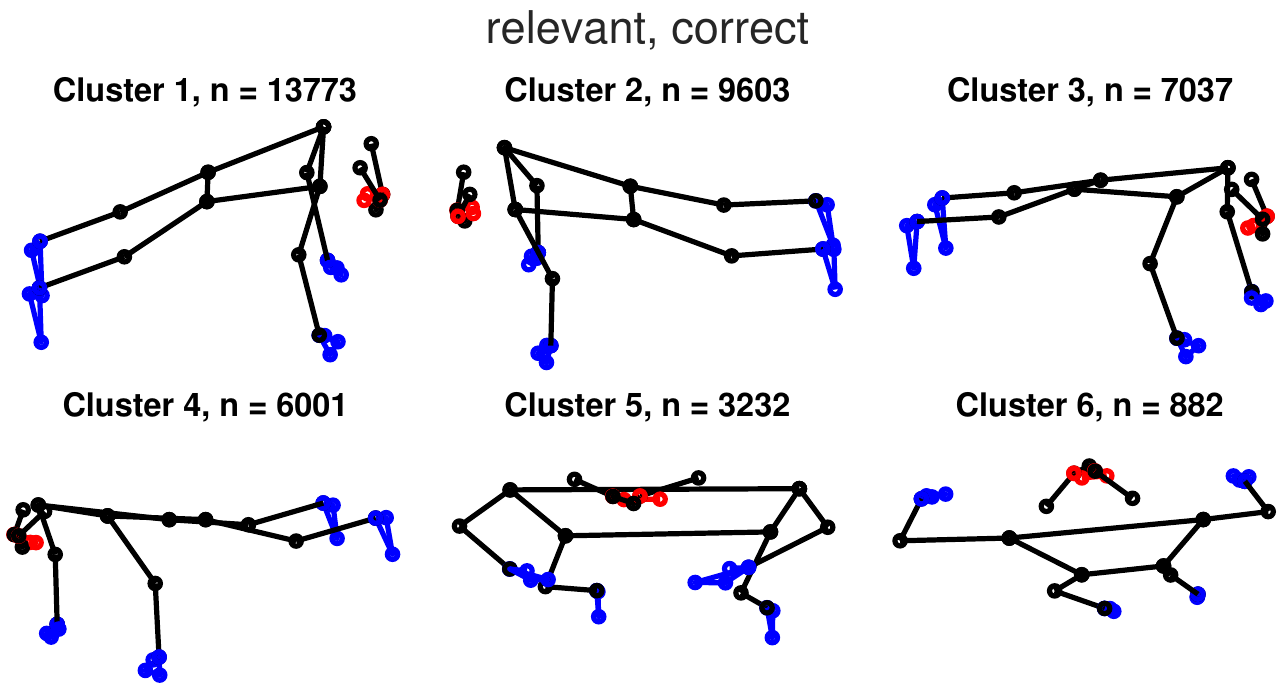}}
	\vspace*{.25cm}
	\centerline{\includegraphics[width=1\columnwidth]{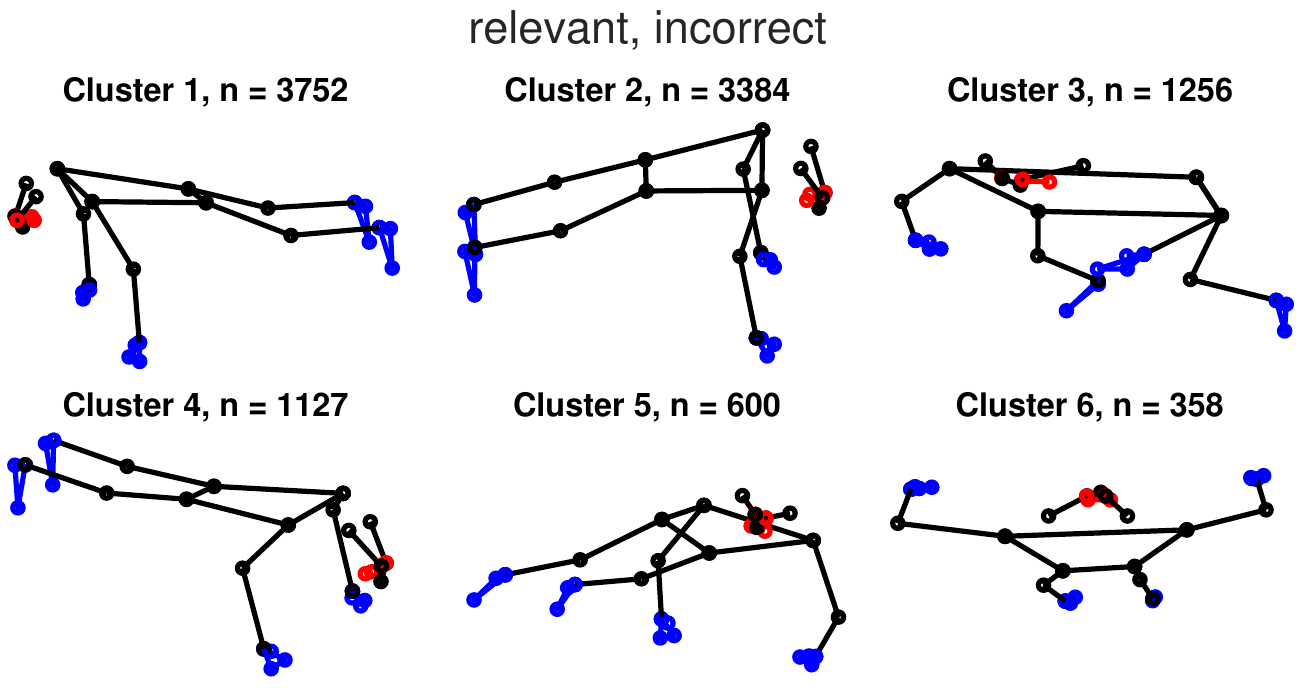}}
	\vspace*{.25cm}
	\centerline{\includegraphics[width=1\columnwidth]{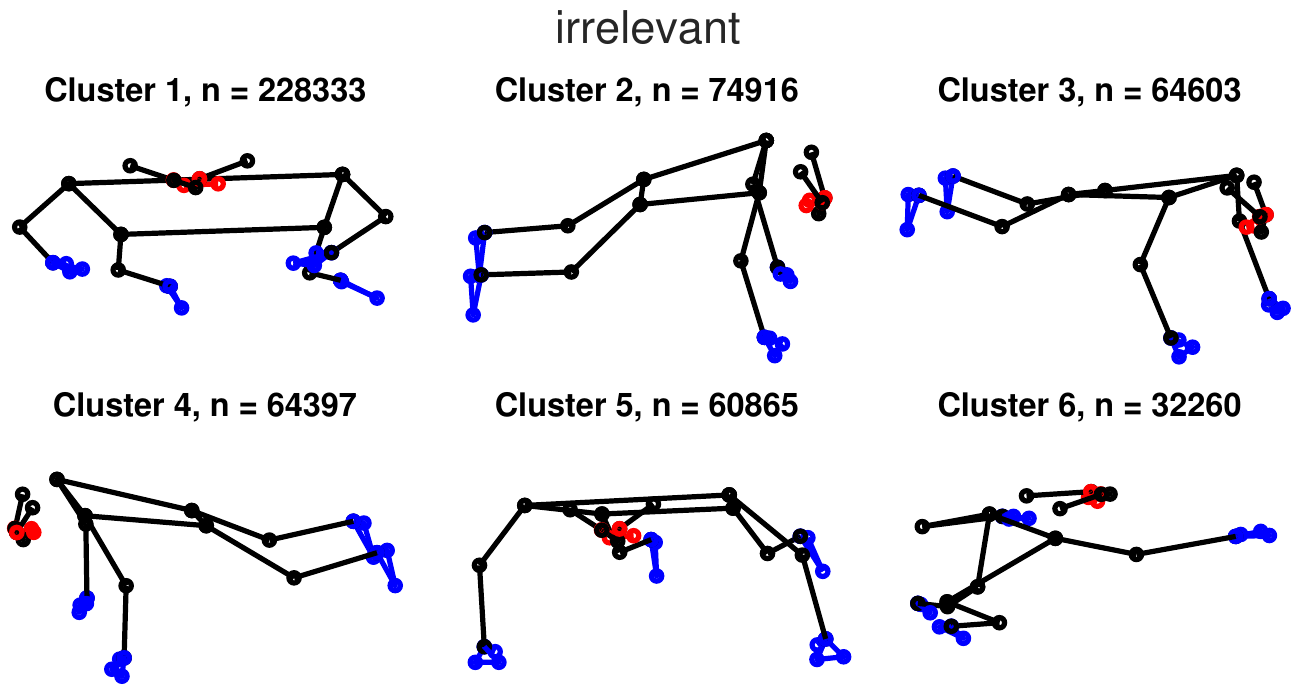}}
	\caption{Visualization of the centroids of the six clusters when clustering ``relevant, correct'', ``relevant, incorrect'', and ``irrelevant'' separately.}
	\label{fig:all_clust} 
	\vspace*{-.2cm} 
\end{figure}

To analyze the datasets defined by NLP separately, Fig.~\ref{fig:all_clust} shows the results of clustering the frames of \emph{each identified class separately} into six clusters.
The four largest clusters of the group \textit{relevant, correct} show people in a regular push-up position. For the group \textit{relevant, incorrect}, the clusters 1, 2, and 4 shows this as well, whereas cluster 5 seems to show some sort of asymmetric positioning. Finally, the largest cluster of the \textit{irrelevant} group seems to show a person facing the camera. Interestingly, the cluster 2 to 5 also seem to show some sort of push-up position.

\begin{figure}[!h] 
	\centerline{\includegraphics[width=.8\columnwidth]{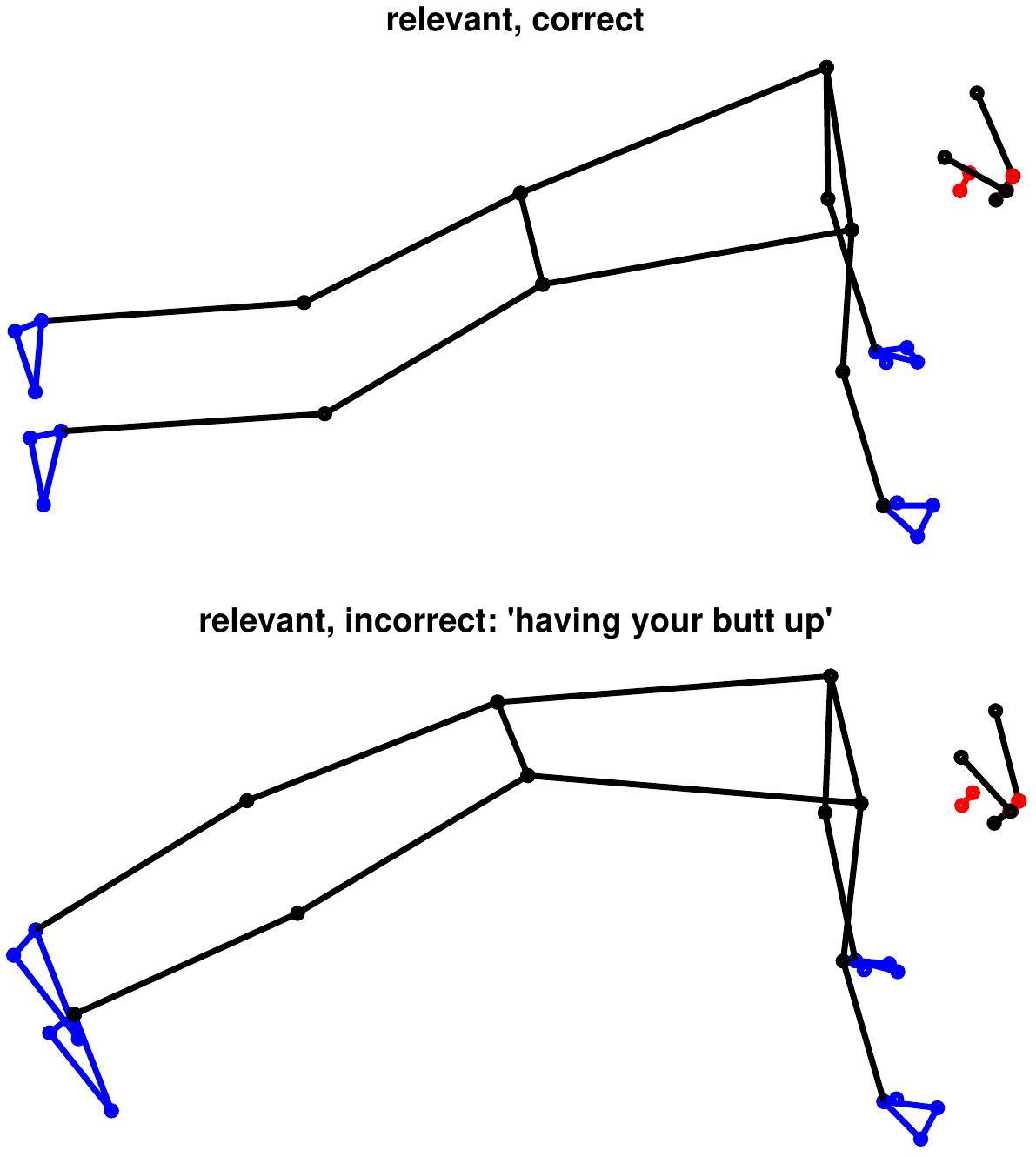}}
	\caption{First row: clip classified as `relevant, correct''. Second row: clip classified ``relevant, incorrect'' (second row) and summarized as ``having your butt up''. Frames were manually selected from the video stream for visualization.}
	\label{fig:correct_incorrect} 
\end{figure}

Using the straight-forward clustering analysis, all temporal information of the frames is lost. Hence, it is not surprising that the clustering algorithm could not pinpoint obvious differences. However, analyzing the labeled clips manually, marked differences could be made out. One example is shown in Fig.~\ref{fig:correct_incorrect}. In the first row, a frame of a clip deemed as \textit{relevant, correct} of one video is shown. In the second row, a frame  of the same video but from a clip classified as \textit{relevant, incorrect} with the NLP-generated summary ``having your butt up'' is shown. Comparing both frames, the summary accurately describes the incorrect pose.

\section{CONCLUSION AND OUTLOOK}
Our study shows that using NLP on social video platforms like YouTube has immense potential when it comes to generating labeled data for video-based human pose analysis. Our k-means clustering algorithm shows major differences in the detected classes. While there are some limitations due to the imprecise time information on the subtitle annotations as well as a correctness classification algorithm that seems to be not complex enough, our results clearly indicate that the NLP classification is worth pursuing. The first steps for future work would be to implement a more complex clustering algorithm, to employ domain knowledge, and more complex motion analysis to better estimate the offset between video and subtitles. Also, we plan to include stronger language models with pre-training on other texts, as well as to examine the gain of using word embeddings for the (A)KW as proposed in~\cite{rudkowsky}. Using biomechanical models for kinematic data analyses could be another potential approach to detect an incorrect pose (or a future malfunction in movement tasks) and amend the classification quality. For example, in~\cite{sharbafi20183d}, such abstract models could distinguish between stroke and unimpaired gaits. Overall, our approach of segmenting and classifying YouTube videos as relevant and irrelevant and labeling them solely based on their subtitles seems very promising for data acquisition and should be further explored in future research. 

\bibliographystyle{IEEEtran}
\bibliography{references}

\end{document}